\documentclass[10pt,twocolumn,letterpaper]{article}
\usepackage[pagenumbers]{wacv} % To force page numbers, e.g. for an arXiv version

\usepackage{graphicx}
\usepackage{amsmath}
\usepackage{amssymb}
\usepackage{booktabs}
\usepackage{balance}

\usepackage[pagebackref,breaklinks,colorlinks]{hyperref}

\usepackage[capitalize]{cleveref}
\crefname{section}{Sec.}{Secs.}
\Crefname{section}{Section}{Sections}
\Crefname{table}{Table}{Tables}
\crefname{table}{Tab.}{Tabs.}

\begin{document}

\title{SVS-GAN: Leveraging GANs for Semantic Video Synthesis}

\author{Khaled M. Seyam\\
Institute of Signal Processing and System Theory, University of Stuttgart, Germany\\
{\tt\small khaled.seyam@iss.uni-stuttgart.de}
\and
Julian Wiederer\\
Mercedes-Benz AG, Stuttgart, Germany\\
{\tt\small julian.wiederer@mercedes-benz.com}
\and
Markus Braun\\
Mercedes-Benz AG, Stuttgart, Germany\\
{\tt\small markus.ma.braun@mercedes-benz.com}
\and
Bin Yang\\
Institute of Signal Processing and System Theory, University of Stuttgart, Germany\\
{\tt\small bin.yang@iss.uni-stuttgart.de}
}
\maketitle

%%%%%%%%% ABSTRACT
\begin{abstract}
In recent years, there has been a growing interest in Semantic Image Synthesis (SIS) through the use of Generative Adversarial Networks (GANs) and diffusion models. This field has seen innovations such as the implementation of specialized loss functions tailored for this task, diverging from the more general approaches in Image-to-Image (I2I) translation. While the concept of Semantic Video Synthesis (SVS)—the generation of temporally coherent, realistic sequences of images from semantic maps—is newly formalized in this paper, some existing methods have already explored aspects of this field. Most of these approaches rely on generic loss functions designed for video-to-video translation or require additional data to achieve temporal coherence. In this paper, we introduce the SVS-GAN, a framework specifically designed for SVS, featuring a custom architecture and loss functions. Our approach includes a triple-pyramid generator that utilizes SPADE blocks. Additionally, we employ a U-Net-based network for the image discriminator, which performs semantic segmentation for the OASIS loss. Through this combination of tailored architecture and objective engineering, our framework aims to bridge the existing gap between SIS and SVS, outperforming current state-of-the-art models on datasets like Cityscapes and KITTI-360.
\end{abstract}

%%%%%%%%% BODY TEXT
\section{Introduction}
\label{sec:intro}

\begin{figure}[t]
  \centering
    \includegraphics[width=0.5\textwidth]{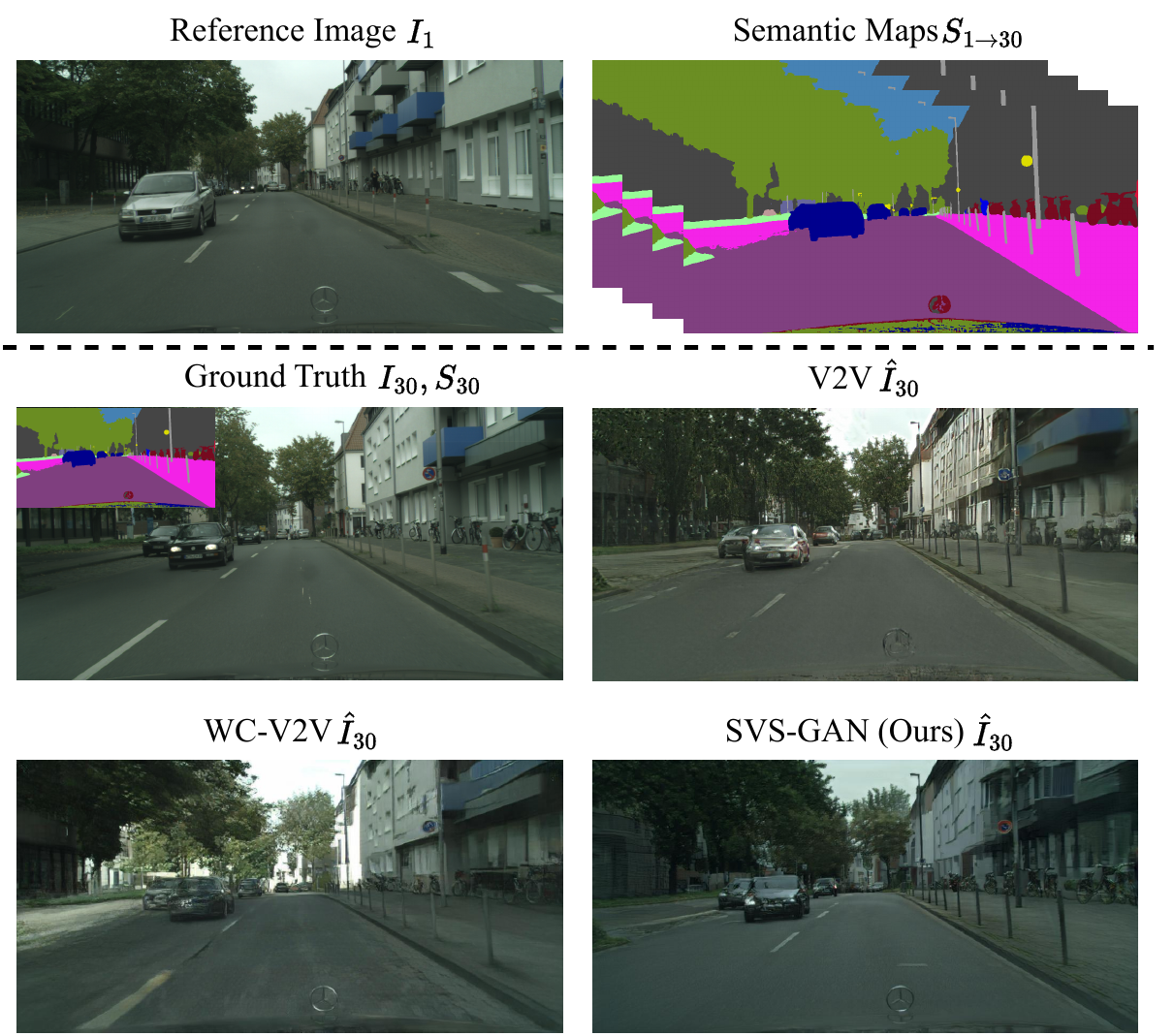}
    \caption{Illustration of our SVS-GAN’s capabilities in generating high-fidelity videos that closely mimic the details of the initial reference image while following the given sequence of semantic maps, demonstrating the model’s effectiveness in accurately synthesizing the 30\textsuperscript{th} frame within the sequence.}

   \label{fig:Intro}
\end{figure}

% Video generation has been gaining more and more attention for the past years, however, advancements in this domain has shown many challenges especially working on huge amount of 3D data compared to 2D images. This is mainly due to the limited hardware capabilities. With the rise of many discriminatory models that work directly on videos for tasks like 3D object detection, there must be more focus on video generative models also. With this, we should be able to augment the data and produce controlled outputs to new scenarios.

% Video generation could be mainly divided into two parts: unconditional, and conditional generation. The former works with generating videos from noise, while in the latter the generation depends on a kind of input to control the process. This includes text, object bounding boxes, or scene layouts. SVS is the application of creating temporal-coherent videos from a sequence of semantic maps. This could be considered as a more challenging side of the SIS task, where the goal is to produce an image from a single semantic map. The main difference is that in the case of SVS, temporal coherence has to be maintained while taking into consideration the limited memory of the GPU. SVS is under conditional generation and offers per pixel control of the produced video. This offers much greater control on the layout of the video compared to text.
Video generation has been drawing more attention in recent years using GANs \cite{GANs,MocoGAN} or diffusion models \cite{Diff, DiffWalt}, but it faces increasing challenges since it involves processing much larger amounts of data in comparison with research areas working with non-temporal/single-shot 2D images. Despite these challenges, there have been significant developments in models designed for perception applications\cite{Bevfusion,3dobjdet,3dsemanticseg,3dsemanticseg2}, like autonomous driving, which requires extensive video training data along with their ground truth labels. This progress highlights the growing importance of video generative models, particularly in the area of controllable generation.  Approaches within this area allow for precise placement and manipulation of generated content to meet specific requirements, which is crucial for creating realistic simulations, customizing outputs for specific applications, and ensuring consistency across generated sequences. Improving these models would enable researchers to augment the data available for training and testing, as well as generate controlled outputs for new and diverse scenarios, such as relevant but rare corner cases in autonomous driving and data generation under domain shifts.

Video generation techniques are divided into two main approaches: unconditional\cite{styleGANV} and conditional\cite{condVideoGen}. Unconditional video generation creates videos from random noise without any guiding input, while conditional video generation relies on specific inputs to direct the creation process. These inputs can range from text descriptions and object bounding boxes to detailed scene layouts, which help in dictating the structure and content of the generated video.

% In this paper, we focus on a type of conditional video generation that uses a sequence of semantic maps as its condition. We term this as Semantic Video Synthesis (SVS), where the objective is to generate a realistic, temporally coherent video that aligns accurately with the sequence of semantic maps. This is a considerably more complex challenge than Semantic Image Synthesis (SIS), which focuses on generating a single static image from a semantic map. The key challenge in SVS lies in maintaining a consistent visual flow across frames, which is critical for creating realistic and useful video sequences. This task is further complicated by the limited memory and processing power of GPUs, which decreases the ability to handle detailed and prolonged video sequences. Nonetheless, SVS offers per-pixel control over the video output, allowing to finely tune every aspect of the video’s appearance and layout. This level of control surpasses what can typically be achieved through textual inputs alone, making SVS a powerful tool for video creation in scenarios requiring high content and geometry customization.

In this paper, we introduce Semantic Video Synthesis (SVS), a form of conditional video generation using semantic map sequences to produce realistic, temporally coherent videos that align accurately with the sequence of semantic maps. This task poses greater complexity than Semantic Image Synthesis (SIS), which generates single images from semantic maps. SVS's primary challenge is maintaining visual consistency across frames, crucial for realistic outputs. However, GPU limitations restrict handling detailed, long video sequences. Despite these challenges, SVS provides per-pixel control over video outputs, offering better precision in video appearance and layout compared to textual inputs, thus enhancing content and geometry customization in video creation.

Despite the fact that SVS has not explicitly been defined as a research area before, the following generic methods already tackle it either using extra modalities along the semantic map or without focusing explicitly on semantic maps as conditional input. The first approach, a Video-to-Video (V2V) synthesis method \cite{vid2vid}, follows a general strategy for video generation similar to the Image-to-Image (I2I) translation framework used in pix2pix \cite{pix2pix}. This approach does not utilize tailored losses or architecture modifications that leverage semantic inputs, leading to poor performance since it fails to capitalize on the rich structural data within the maps. The second study, World-Consistent Video-to-Video (WC-V2V)\cite{WC-vid2vid}, enhances the generator architecture by incorporating SPADE\cite{SPADE} blocks and focuses on using semantic maps as input. However, it also integrates additional 3D world inputs, thereby deviating from the primary goal of SVS, which is to generate realistic videos directly from semantic map sequences without additional data.

To address the mentioned limitations and motivated by the advancements in SIS, we introduce SVS-GAN, a framework designed to bridge the gap between SVS and SIS. Figure \ref{fig:Intro} showcases results from our model, which processes a sequence of semantic maps alongside a reference image \(I_1\)—either real or generated by any SIS framework. SVS-GAN then autoregressively generates the subsequent sequence of frames.
% A detailed view of our complete architecture is provided in figure \ref{fig:Generator}, featuring our generator \(G\), the per-image discriminator \(D_I\), and the video discriminator \(D_V\).

% As seen in figure \ref{fig:Intro}, our framework expects a reference image $I_1$, that could be either real or generated using a SIS framework, and a list of semantic maps where our framework generates from it 

% Similar to V2V and WC-V2V, our framework consists of a next frame prediction generator $G$, an image discriminator $D_I$, and a video discriminator $D_V$. The auto-regressive nature of our generator enables the generation of extended video sequences. $G$ accepts the previously generated image, previous semantic map, and current semantic map as inputs. The idea is to produce temporally coherent videos that maintain style and object consistency. 

Our framework introduces a novel triple pyramid generator architecture tailored specifically for SVS. Additionally, we are the first to introduce the OASIS\cite{OASIS} loss in video generation, which significantly improves semantic alignment metrics. This is facilitated by using our encoder-decoder discriminator, which is capable of segmenting images. Our evaluations demonstrate that SVS-GAN surpasses existing state-of-the-art approaches in image quality, temporal coherence, and semantic alignment without requiring additional inputs and by considering only one previous frame. Additionally, our model operates efficiently on a single GPU at a resolution of $1024\times512$, achieving an inference time of 40ms per frame. 

Our architecture is versatile enough to be applied to any SVS task, yet we have specifically targeted driving scenarios due to their complexity and growing relevance in the field of autonomous driving. These scenarios are particularly challenging because they involve dual movement dynamics: the motion of the camera, represented by the ego vehicle, and the interactions of other vehicles and pedestrians. We validate our framework on the Cityscapes Sequence\cite{Cityscapes} and KITTI-360\cite{Kitti360} datasets, achieving state-of-the-art performance on both. A detailed ablation study highlights the impact of each component in our framework.

\section{Related Work}
\label{sec:related}
Data generation plays a crucial role in the field of machine learning, serving as a data augmentor for training various discriminative models. This includes a wide array of data types like images, videos, and additional sensor outputs such as LiDAR. Within neural networks, two main approaches stand out for data generation: GANs and diffusion models. Recently, diffusion models have made significant improvements in producing realistic outputs, surpassing GANs in some aspects. However, they are hampered by long training and inference times, and in scenarios such as SIS, their output quality remains comparable to that of GANs\cite{GANSIS}.

\subsection{Semantic Image Synthesis}

SIS falls under the broader category of I2I translation, where the conditioning image is a semantic map. The concept of I2I using GANs was pioneered by the pix2pix framework, which introduced an approach to translate images from one domain $X$ to another domain $Y$ without specific domain constraints\cite{pix2pix}. This framework could perform various tasks, such as image colorization and SIS. Subsequent developments in I2I mainly focused on application-specific enhancements. Several studies, including those on SPADE\cite{SPADE}, OASIS\cite{OASIS}, and DP-GAN\cite{DPGAN}, have concentrated solely on SIS, evolving architectures and loss functions tailored to generate realistic images that closely align with the semantic map. SIS typically functions within a supervised learning framework, relying on paired datasets. However, the scarcity of such data has encouraged interest in Unsupervised SIS (USIS)\cite{usis} and semi-supervised approaches\cite{sssis}.

SPADE\cite{SPADE} emerged as a solution in SIS by integrating a spatially adaptive denormalization layer into the generator. The authors demonstrated that directly feeding the semantic map to the generator was not optimal. Instead, SPADE modifies the normalization process by utilizing segmentation maps to adaptively scale and shift the normalized output. This technique enables the generation of images that adhere more faithfully to the input semantic layout, preserving spatial information typically lost in standard normalization layers.

Previously, many frameworks presented the discriminator with the real or fake image alongside its corresponding semantic map. However, this approach did not guarantee that the discriminator would leverage the semantic map as a guide in the discrimination process. OASIS\cite{OASIS} addressed this by implementing an encoder-decoder architecture in the discriminator, that not only produces real/fake prediction, but segments the image. This structure encourages the discriminator to learn image segmentation, while the generator tries to fool the discriminator by creating images that can be accurately segmented by it.

\subsection{Video-to-Video}

The definition of V2V synthesis has varied across studies, with some interpreting it as a method for converting between any two video domains\cite{vid2vid}, and others specifically linking it to SVS\cite{WC-vid2vid}. In our study, we will use V2V to denote the transformation of a video from one arbitrary domain $X$ to another $Y$. Similarly, we will define SVS in a manner similar to SIS, with the domain $X$ limited to sequences of semantic maps.

The initial application of GANs to these tasks was in 2018\cite{vid2vid}, which explored video transformations such as from pose to dance, from sketches to faces, and SVS. The generator $G$ was designed to consider both the two previous frames from the source and target domains, along with the current source image, to produce the next frame in the target sequence. The work presented in \cite{vid2vid} approached the task as an autoregressive next-frame prediction, incorporating a specialized module to calculate optical flow\cite{OpticalFlow} for image warping, complemented by an additional module for generating rapidly-moving foreground objects that the optical flow could not track. Subsequently, Fast-V2V\cite{fast_vi2vid} was introduced, adding to the system spatial-temporal compression techniques to accelerate inference and minimize GPU load. Additionally, the scarcity of paired datasets motivated the development of unsupervised frameworks\cite{unsupv2v}.

A later development in this field was the introduction of WC-V2V, which prioritized maintaining consistency across the entire world in the video rather than ensuring only frame-to-frame continuity\cite{WC-vid2vid}. Unlike traditional V2V that processes just the two preceding frames, WC-V2V utilizes the entirety of preceding frames to synthesize the current frame, necessitating a 3D world representation derived from actual images using structure from motion (SfM)\cite{SFM1, SFM2}. This requirement, however, limits its applicability to pure SVS, as creating a reliable 3D model from real images can be impractical.

% Our research introduces an approach aimed exclusively at pure SVS. We utilize a sequence of semantic maps as inputs and generate a corresponding sequence of realistic spatial-temporal coherent images. This is achieved through a tailored architecture and custom-designed loss functions, designed to ensure high fidelity, maintain temporal coherence, and accurately adhere to the semantic maps in the synthesized video sequences. 

Our main contributions are threefold:
\begin{enumerate}
    \item We formally define SVS and introduce a dedicated framework for pure SVS. Utilizing a sequence of semantic maps as inputs, our framework generates a corresponding sequence of realistic, and spatially and temporally coherent video frames.
    \item We develop a novel triple-pyramid generator architecture that leverages semantic maps and information from the past frame to predict visually coherent and semantically accurate frames. 
    \item We integrate the OASIS loss \cite{OASIS} into our SVS framework, enhancing the alignment of synthesized videos with their corresponding semantic maps.

\end{enumerate}

Our contributions have led to state-of-the-art performance on datasets such as Cityscapes and KITTI-360, excelling in metrics that assess image quality, video quality, and adherence to the semantic map.

% Our main contributions are threefold:
% \begin{enumerate}
%     \item We introduce SVS-GAN, a novel framework specifically tailored for SVS, which leverages a novel triple-pyramid generator architecture to generate temporally coherent video sequences from sequences of semantic maps.
%     \item Our approach integrate the OASIS loss in conjunction with SPADE blocks within the generative model, significantly enhancing the semantic alignment and realism of the synthesized videos compared to existing methods.
%     \item We provide extensive evaluations of our model on challenging datasets, including Cityscapes and KITTI-360, demonstrating superior performance in terms of image quality, temporal coherence, and semantic accuracy.
% \end{enumerate}

% Our main contributions are threefold:
% \begin{enumerate}
%     \item We define and introduce a targeted approach for pure Semantic Video Synthesis (SVS). Utilizing a sequence of semantic maps as inputs, our framework generates a corresponding sequence of realistic, spatial-temporally coherent images.
%     \item We develop a novel triple-pyramid generator architecture capable of producing the next frame based on the past frame and both past and current semantic maps, enhancing the temporal continuity and visual coherence of the generated video.
%     \item We integrate the OASIS loss \cite{OASIS} into SVS, significantly improving the fidelity and realism of the synthesized videos. This advancement has led to state-of-the-art performance on challenging datasets such as Cityscapes and KITTI-360.
% \end{enumerate}

\section{Method}
\label{sec:method}

\begin{figure*}
\centering
\includegraphics[width=1.0\textwidth]{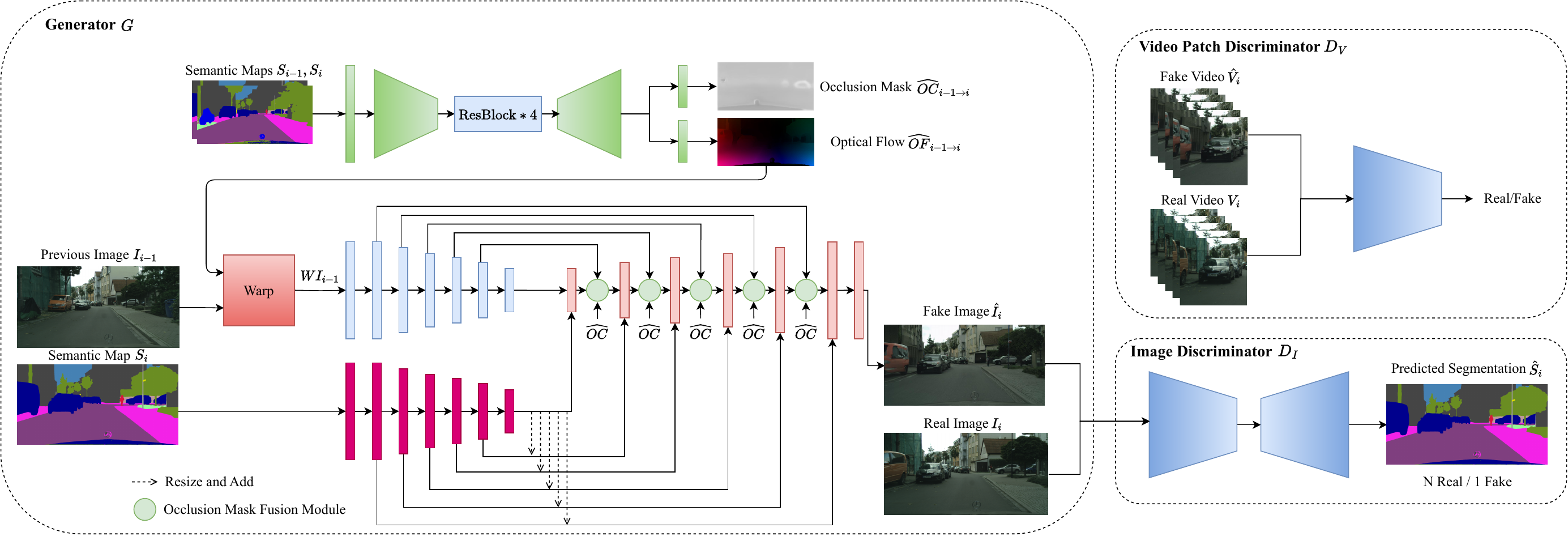}
  % \caption{
  % % Our generator architecture where we can see the optical flow and the occlusion map network. Bottom we can see our triple-pyramid architecture with the warped image encoder, spatially adaptation encoder, and our image generation decoder.
  % Illustration of our architecture showcasing our generator $G$ and discriminator $D_I$ and $D_V$. In $G$ we can see the optical flow and the occlusion map network on top. Below, the triple-pyramid architecture is shown, featuring the warped image encoder, the semantic map encoder, and the image generation decoder. On the right we can see the encoder/decoder $D_I$ which segments the image to use the OASIS loss. and above it is the video patch discriminator. 
  % }

\caption{Architectural illustration highlighting the generator \( G \) and the discriminators \( D_I \) and \( D_V \). The generator features an optical flow and occlusion map network illustrated at the top, followed by a triple-pyramid structure comprising a warped image encoder, a semantic map encoder, and an image generation decoder. The decoder employs SPADE blocks to incorporate the semantic map and utilizes \(\widehat{OC}_{i-1\rightarrow i}\) to modulate features passed through the skip connections using the occlusion mask fusion module. On the right, the discriminator \( D_I \) employs an encoder/decoder setup for image segmentation using the OASIS loss, while above, \( D_V \) functions as a video patch discriminator, aiming for temporal consistency.}

  \label{fig:Generator}
\end{figure*}

% Our problem could be formulated as next frame prediction given the current semantic map. We perform the video generation in an auto-regressive manner where each frame depends appearance wise on the previously generated frame $F_{i-1}$ and follow the current semantic map $S_i$. This gives us the flexibility to generate variable video lengths. 

% Our architecture consists of 3 main components: (i) Generator $G$ that accepts the past and current semantic maps ($S_{i-1}$, $S_i$) to assess the movement and calculate the optical flow. Moreover, $G$ uses $F_{i-1}$ to extract the appearance from the previous image. Lastly, it uses $S_i$ to guide the generator to follow this semantic map. (ii) Image Discriminator $D_I$ only looks at one image per time to assess how realistic the image looks as a whole and guides the generator to produce results following the semantic map. (iii) Video Discriminator $D_V$ that takes multiple frames at once to assess the temporal coherence between them.

Our model addresses the task of SVS by predicting the next frame based on a given semantic map through an autoregressive video generation process. Each frame $I_i$ depends on the previously generated image, denoted as $I_{i-1}$, and aligns with the current semantic map $S_i$. This approach allows the creation of videos with variable lengths that are tailored to the input semantic maps.

Our new framework for SVS consists of three primary components:\begin{itemize}
    \item \textbf{Generator ($G$)}: The generator takes as input both past and current semantic maps ($S_{i-1}$, $S_i$) to calculate the optical flow. It also utilizes appearance information from the previous image $I_{i-1}$, and uses the current map $S_i$ to ensure the generated frame adheres to the specified semantic guidance.
    \item \textbf{Image Discriminator ($D_I$)}: This component evaluates individual frames independently to determine their realism and conformity to the semantic map.
    \item \textbf{Video Discriminator ($D_V$)}: The video discriminator assesses groups of generated frames together to evaluate their temporal and spatial coherence. This ensures the sequence of generated frames is smooth and transitions are logically consistent, contributing to a realistic video sequence.
\end{itemize}

% To generate frame $F_i$ we expect the past and current semantic maps ($S_{i-1}$, $S_i$) to assess the movement and calculate the optical flow. Moreover, we use $F_{i-1}$ to extract the appearance from the previous image. Lastly, we use $S_i$ to guide the generator to follow this semantic map. 

\subsection{Generator}

Our generator architecture is designed to be efficient while producing realistic results, utilizing about one-fourth of the parameters of frameworks like V2V\cite{vid2vid}. It contains two primary segments: an optical flow predictor and a triple-pyramid network for next frame prediction. While our generator architecture incorporates an optical flow network similar to that used in V2V, our method diverges by integrating optical flow and occlusion maps not merely for warping the previous frame and blending it with the generated image, but by embedding these elements into deeper layers of the frame synthesis process for enhanced generation. The optical flow predictor functions as an encoder-decoder network incorporating ResNet\cite{Resnet} blocks to process inputs ($S_{i-1}$, $S_i$). This network generates the predicted optical flow $OF_{i-1 \rightarrow i}$ and an occlusion map $OC_{i-1 \rightarrow i}$. Here, $OF_{i-1 \rightarrow i}$ depicts the movement of each pixel from $I_{i-1}$ to $I_i$, while $OC_{i-1 \rightarrow i}$ identifies areas potentially occluded in the transition and requiring regeneration.

% Utilizing $OF_{i-1 \rightarrow i}$, the previous image \( I_{i-1} \) is warped to produce \( WI_{i-1} \) by shifting each pixel according to its corresponding motion vector in \( OF_{i-1 \rightarrow i} \). This warped image feeds into the first pyramid of our network, an image encoder, designed to extract image features for constructing the current image $I_i$ in a U-Net\cite{unet} configuration with skip connections (refer to Figure \ref{fig:Generator}). $OC_{i-1 \rightarrow i}$ modulates these skip connections, guiding the balance between generated content and information carried over from $WI_{i-1}$.

Utilizing \( OF_{i-1 \rightarrow i} \), the previous image \( I_{i-1} \) is warped to produce the warped image \( WI_{i-1} \) by shifting each pixel according to its corresponding motion vector in \( OF_{i-1 \rightarrow i} \). This warped image feeds into the first pyramid of our network, an image encoder, designed to extract image features for constructing the current image \( I_i \) in a U-Net\cite{unet} configuration with skip connections (refer to Figure \ref{fig:Generator}). The modulation of these skip connections by \( OC_{i-1 \rightarrow i} \) is used to determine the balance between the generated content and the information carried over from \( WI_{i-1} \). \( OC_{i-1 \rightarrow i} \), with pixel values ranging from 0 to 1, indicates whether a pixel should feature newly generated content or retain features from the warped previous image. This dynamic scaling optimizes the integration of these inputs to construct \( I_i \), thereby effectively managing the blend based on pixel-specific occlusion.

% Features from \( WI_{i-1} \) are scaled by \( 1 - OC_{i-1 \rightarrow i} \) and the new image features by \( OC_{i-1 \rightarrow i} \).

The second pyramid, another encoder, accepts  $S_i$ as input and produces per-level features for the SPADE\cite{SPADE} blocks within the generator's decoder. Similar to DP-GAN\cite{DPGAN}, the bottleneck features, which provide the largest field of view, are concatenated to the features across all levels to maintain a global perspective. This configuration enables more accurate denormalization by integrating global context and detailed local features, thereby improving the fidelity and consistency of the generated images relative to the input semantic maps.

The third pyramid, functioning as the decoder, employs multiple SPADE ResNet blocks. These blocks utilize the features extracted from $WI_{i-1}$ and $S_i$ to generate the current image $I_i$, ensuring fidelity to the appearance of $I_{i-1}$ and adherence to the semantic guidance from $S_i$.

\subsection{Discriminators}
% The goal of the discriminator is to assure spatio-temporal coherence and adherence to the semantic map. This is best done using two discriminators where $D_I$ focuses on the image level and following the semantic map while $D_V$ focuses on the video level and assuring temporal coherence. 

% We chose $D_I$ to be an encoder-decoder network that uses the OASIS loss to guide the generator. The input to our discriminator is either the real image $I_i$ or generated image $\hat{I}_i$. In the case of $I_i$, $D_I$ learns to segment the image to the correct classes. Otherwise, it tries to predict an extra class which represents the fake class. This makes sure that the generator generates images that are both spatially realistic and closely follow $S_i$. For $D_V$ we use the same network as V2V which is a patchwise discriminator to assure temporal coherence.

The primary objective of the discriminator is to ensure spatial-temporal coherence and fidelity to the semantic map. This is achieved through the use of two discriminators: $D_I$ and $D_V$. The image discriminator $D_I$ concentrates on validating each image's realism and adherence to the semantic map, while the video discriminator $D_V$ ensures temporal coherence across a video sequence.

$D_I$ is structured as an encoder-decoder network similar to DP-GAN\cite{DPGAN}, utilizing the OASIS loss to effectively guide the generator. It accepts as input either a real image $I_i$ or a generated image $\hat{I}_i$ and produces a predicted segmentation map $\hat{S}_i$. In the case of $I_i$, $D_I$ is tasked with segmenting the image into correct semantic labels. Conversely, when evaluating $\hat{I}_i$, it attempts to identify an additional class indicative of the 'fake' class. This segmentation-based loss specifically provides detailed, pixel-level feedback, enhancing the semantic alignment of the generated images with the input map $S_i$. This loss-driven feedback forces the generator to refine its output to closely match $S_i$.

For \(D_V\), we employ a network model similar to V2V\cite{vid2vid}, which functions as a patchwise discriminator to check temporal coherence. In our framework, we utilize two video discriminators, with each one targeting a different temporal scale to ensure coverage of longer temporal sequences. This means each discriminator analyzes frames at varying rates to assess broader temporal coherence. Each \(D_V\) examines three frames simultaneously to evaluate their temporal alignment, thereby enhancing the overall realism of the video.

\subsection{Losses}
% Many losses are used to optimize our architecture. We use a weighted cross entropy loss for the result of $D_I$ which focuses on under represented classes. We use another adversarial GAN loss with the result of $D_V$. Beside those, we use a VGG loss to assure that the generated images are close to the real ones. We also use a feature matching loss based on the extraced features of the decoder of $D_I$. Lastly, we use a flow and a warping loss to motivate the optical flow calculating network to produce accurate results.

Our model architecture employs a comprehensive suite of loss functions, each designed to optimize specific characteristics of the generated outputs. These loss functions are briefly summarized below, with detailed formulations and additional discussions available in the appendix.

\begin{itemize}
% \item \textbf{OASIS Adversarial Loss $\mathcal{L}_{D_I}$ and $\mathcal{L}_{G_I}$~\cite{OASIS}:} The discriminator's loss, $L_{D_I}$, treats the task as a semantic segmentation problem, adding a class for generated (fake) images. $D_i$ aims to segment real images $I_i$ into $S_i$ and identifies generated images $\hat{I}_i$ by assigning each pixel to the fake class. The generator's loss, $L_{G_I}$, encourages $G$ to produce images that not only are classified by $D_I$ as real but also accurately segmented into their correct semantic classes $S_i$, thereby closely adhering $\hat{I}_i$ to the semantic map. Both losses utilize cross-entropy between $S_i$ and the predicted $\hat{S}_i$ to enhance both semantic accuracy and realism.

% \item \textbf{OASIS Adversarial Loss $\mathcal{L}_{D_I}$ and $\mathcal{L}_{G_I}$~\cite{OASIS}:} $D_I$ functions as a semantic segmentation network, producing a semantic map $\hat{S}_i$. The loss for the discriminator, $L_{D_I}$, uses a cross-entropy loss to encourage accurate segmentation of real images $I_i$ by comparing the predicted map $\hat{S}_i$ with the $S_i$. For generated images $\hat{I}_i$, $D_i$ is encouraged to assign each pixel to an additional class designated for the fake label. Concurrently, the generator's loss, $L_{G_I}$, aims to deceive $D_I$ by generating images $\hat{I}_i$ that are misclassified as real and correctly mapped to their respective semantic classes $S_i$, thereby aligning them closely with the intended semantic maps. This interaction enhances both the realism and semantic accuracy of the images produced.

\item \textbf{OASIS Adversarial Loss $\mathcal{L}_{D_I}$ and $\mathcal{L}_{G_I}$~\cite{OASIS}:} $D_I$ acts as a semantic segmentation network, generating a semantic map $\hat{S}_i$. The discriminator's loss, $L_{D_I}$, uses cross-entropy to ensure precise segmentation of real images $I_i$ by comparing $\hat{S}_i$ with $S_i$. For generated images $\hat{I}_i$, $D_I$ aims to assign each pixel to an additional fake label class. Meanwhile, the generator's loss, $L_{G_I}$, is designed to deceive $D_I$ by producing images $\hat{I}_i$ that are misclassified as real and accurately mapped to their corresponding semantic classes $S_i$, using also cross-entropy, therefore enhancing the realism and semantic accuracy of the generated images.

    \item \textbf{Adversarial Loss $\mathcal{L}_{adv}$~\cite{vid2vid,WC-vid2vid}:} Applied to video sequences, this loss utilizes a discriminator to distinguish between real and synthetic videos, enforcing that generated videos are indistinguishable from real videos, thus promoting temporal stability and coherence.
    \item \textbf{VGG Loss $\mathcal{L}_{VGG}$~\cite{VGGGan}:} Measures the perceptual difference between the feature maps of generated $\hat{I}_i$ and real images $I_i$ obtained from various layers of a pre-trained VGG network\cite{VGG}, aiming to reduce these discrepancies to enhance visual similarity.
    % \item \textbf{Feature Matching Loss $\mathcal{L}_{FM}$~\cite{fmloss}}: Aligns intermediate representations of real and generated images by minimizing the L1-norm difference between feature maps extracted from multiple layers of the discriminators.

    \item \textbf{Feature Matching Loss $\mathcal{L}_{FM}$~\cite{fmloss}}: Aligns intermediate representations of real and generated images by minimizing the L1-norm difference between feature maps extracted from multiple discriminator layers.
    
\item \textbf{Flow and Warping Loss $L_{Flow}$~\cite{vid2vid,WC-vid2vid}:} This loss function evaluates the accuracy of predicted optical flows by comparing them with pseudo ground truth flows derived from FlowNet2\cite{flownet}. Additionally, it incorporates a warping loss that assesses the consistency of $WI_{i-1}$ with their subsequent frames $I_i$. $L_{Flow}$ aims to minimize motion discrepancies between consecutive frames, enhancing the realism and temporal consistency of the video output.

\end{itemize}

\section{Experiments}
% \textbf{Experimental Setup:} We use a single NVIDIA A6000 GPU for training of our model, this is much less than other frameworks where the number of parameters of our generator is less than 35\% than V2V and WC-V2V. We use progressive learning both in the length of the sequence and the resolution of the frame. We train our model fully with a constant resolution and increase the sequence length the model works with till we reach a sequence of 30 frames. The progressive learning is done by adding an extra SPADE block in the decoder of the generator.

% In the training of the base model, we train for 20 epoch with a learning rate (lr) of 0.0002 starting with a sequence of length 6 and doubles every 5 epochs. We then continue the training for another 20 epochs with a decaying lr. We use Adam optimizer with $\beta_1 = 0.5$, and $\beta_2 = 0.999$.

% \textbf{Experimental Setup:} 
The following describes the implementation details, introduces the datasets and metrics used throughout the experiments, and finally presents the results.
\subsection{Implementation Details} 

% We trained our model using a single NVIDIA A6000 GPU. We employed a progressive learning approach that includes both increasing the sequence length\cite{vid2vid} and the spatial resolution\cite{PL}. Initially, the base model is trained at a constant resolution, with the sequence length progressively increasing up to 30 frames. Spatial progression is facilitated by adding an additional SPADE block to the generator’s decoder, which enhances the model’s ability to handle higher resolutions as training progresses.

% In the base model training phase, we conducted training over 20 epochs with an initial learning rate of 0.0002, starting with a sequence length of 6 frames, which increased to 12, then 24, and finally 30 frames, adjusting every 5 epochs. Subsequently, we continued the training for another 20 epochs with a decaying learning rate. The Adam optimizer\cite{Adam} was used with parameters $\beta_1 = 0.5$ and $\beta_2 = 0.999$. 

We trained our model using a single NVIDIA A6000 GPU and employed a progressive learning approach that includes both increasing the sequence length~\cite{vid2vid} and the spatial resolution~\cite{PL}. The base model is initially trained over 20 epochs at a constant resolution, with the sequence length starting at 6 frames and progressively increasing to 12, 24, and finally 30 frames, with increments every 5 epochs. After completing these 20 epochs, training continues for another 20 epochs with a decaying learning rate. Following this initial 40-epoch training period, an additional SPADE block is integrated into the generator’s decoder to facilitate spatial progression. This enhancement enables the model to handle higher resolutions. Subsequently, the model undergoes further training for an additional 8 epochs with the sequence length progressively increasing. Throughout this process, the Adam optimizer~\cite{Adam} is utilized with parameters $\beta_1 = 0.5$ and $\beta_2 = 0.999$.

\subsection{Datasets}
We trained our model on two driving datasets: Cityscapes Sequence and KITTI-360. These datasets were chosen for their challenging scenarios, which include the movement of the ego vehicle (the camera) as well as the dynamic movement of other objects such as cars and pedestrians. Although our model was specifically trained on these driving datasets, it is designed to generalize to non-driving scenarios as well.

\textbf{Cityscapes Sequence\cite{Cityscapes}:} The Cityscapes Sequence dataset extends the well-known Cityscapes dataset, specifically tailored for video-based urban scene analysis. It encompasses 150,000 high-resolution frames (2048 x 1024 pixels), drawn from video sequences captured across 30 cities at 17 frames per second. Each sequence consists of 30 frames. Notably, ground truth semantic labels are provided only for a single frame in each sequence. To address this limitation, we employed OneFormer\cite{oneformer}, which was trained on the Cityscapes dataset, to produce segmentation and instance maps for all images. Although these segmentations are not fully accurate, this pseudo ground truth is sufficient to enable our model to generate realistic videos.

Our model is trained using the 2,975 training sequences and is assessed on the 500 validation sequences. We start by training our model on resolution $512\times256$ as our base and then add an extra SPADE block to reach resolution $1024\times512$.

% \textbf{KITTI-360 Dataset\cite{Kitti360}:}
% The KITTI-360 dataset builds on the original KITTI\cite{KITTI} dataset by offering an extensive collection of over 80,000 images (1400 x 425 pixels) capturing urban and rural road scenes for autonomous driving research. Annotations include 2D bounding boxes for object detection and semantic labels for images, providing comprehensive ground truth data for both object positioning and scene understanding.

% The dataset features sequences that vary in length, with a minimum of 136 frames and a maximum of 8294 frames, captured at a rate of 11 frames per second. This variety allows for testing models on both short and extended sequences, with initial training on sequences of length 30 and scaling up to resolutions of $1024\times 256$.

 \textbf{KITTI-360 Dataset\cite{Kitti360}:}
The KITTI-360 dataset enhances the original KITTI\cite{KITTI} by providing detailed 3D urban scene perception for autonomous driving research. This dataset comprises over 80,000 images (1400 x 425 pixels) alongside more than 80,000 dense 3D laser scans, covering approximately 75 km of urban and rural roads.

Annotations include 2D and 3D bounding boxes for object detection and semantic labels for both images and 3D point clouds. Data capture involves two color video cameras and two laser scanners.

The sequences in this dataset vary in length, with a minimum sequence length of 136 frames and a maximum of 8294 frames taken at 11 frames per second. This gives us the ability to test our model on long sequences while only trained on sequences of length 30. We start with a resolution $512\times 128$ and reach $1024\times 256$.

\subsection{Metrics}
We evaluate our SVS-GAN using several key metrics to assess both visual quality and semantic accuracy. Fréchet Inception Distance (FID)\cite{FID}, calculated using feature vectors from the Inception-v3 network\cite{inception} trained on ImageNet\cite{imagenet}, quantifies how closely generated images match real images in content and style. Fréchet Video Distance (FVD)\cite{FVD} is used to measure how closely generated videos mimic real ones. Traditionally, FVD employs FVD\textsubscript{i3d} features from the I3D model, which focuses on frame appearance and includes some temporal elements. However, \cite{FVDcd} demonstrates that FVD\textsubscript{cd}, using features from the VideoMAE-v2\cite{videomae} network, provides a more precise assessment of temporal coherence, offering a more accurate evaluation of video realism. Mean Intersection over Union (MIoU) evaluates the semantic accuracy, using predictions from the DRN-D-105 network\cite{drn} to measure the overlap of semantic labels with ground truths. Lower FID and FVD scores, along with higher MIoU values, indicate improved performance of the generated videos. For a comprehensive exploration of these metrics, see the supplementary material.

\begin{figure*}
\centering
\includegraphics[width=1.0\textwidth]{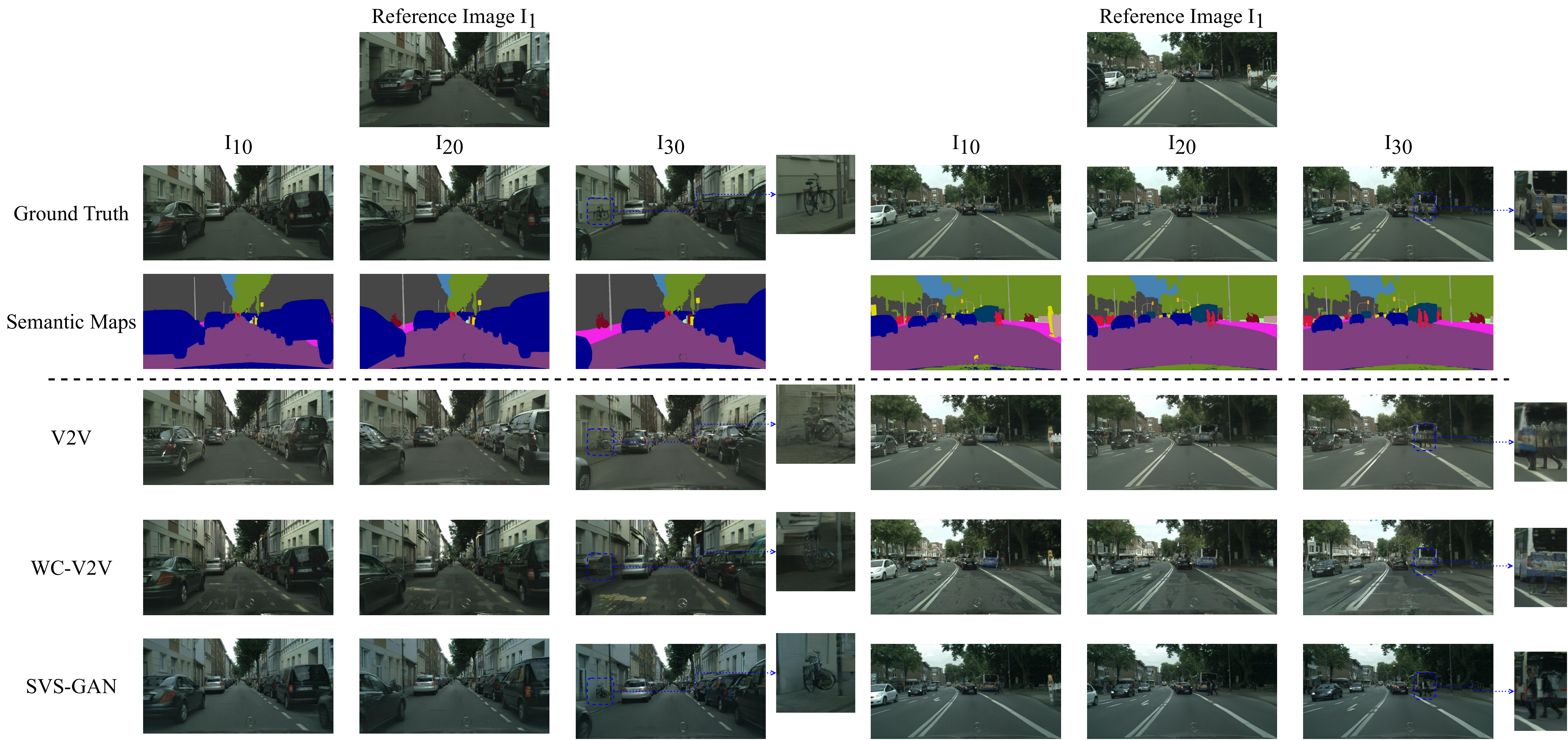}
  \caption{
  Comparison of model performances on the Cityscapes sequence dataset: On the two displayed scenes our model, compared to V2V and WC-V2V models, demonstrates enhanced detail capture, particularly in the road surfaces and vehicles.
  }
  \label{fig:QualitativeCS}
\end{figure*}

% These metrics provide a comprehensive assessment of the synthesized videos, evaluating not only the perceptual quality but also the accuracy of semantic content reproduction.

\subsection{Experimental Results}

% \begin{table}
%   \centering
%   {\small{
%   \begin{tabular}{@{}lccc@{}}
%     \toprule
%     Method & FID (↓) & FVD (↓) & MIoU (↑) \\
%     \midrule
%     V2V & 15.60 & 168.31 & 47.71 \\
%     WC-V2V & 20.63 & 208.22 & 49.75 \\
%     SVS-GAN (Ours) & \textbf{13.98} & \textbf{94.72} & \textbf{56.81} \\
%     \bottomrule
%   \end{tabular}
%   }}
%   \caption{Quantitative results on the Cityscapes sequence dataset.}
%   \label{tab:performance_comparison_CS}
% \end{table}

\begin{table}
  \centering
  {\small{
  \begin{tabular}{@{}lccccc@{}}
    \toprule
    Method & FID ↓ & FVD\textsubscript{i3d} ↓ & FVD\textsubscript{cd} ↓ & MIoU ↑ & FPS\\
    \midrule
    V2V & 69.07 & 126.71 & 97.90 & 55.4  & 4.3\\
    WC-V2V & 49.89 & 127.36 &110.44 & 62.0  & 3.4\\
    Fast-V2V & 89.57 & 322.06 & 223.57 & 35.1  & 24.8\\ 
    SVS-GAN & \textbf{42.34} & \textbf{85.30}  & \textbf{66.91}& \textbf{66.0} & \textbf{25.6} \\
    \bottomrule
  \end{tabular}
  }}
  \caption{Quantitative results on the Cityscapes sequence dataset.}
  \label{tab:performance_comparison_CS}
\end{table}

\begin{table}
  \centering
  {\small{
  \begin{tabular}{@{}lcccc@{}}
    \toprule
    Method & FID (↓) & FVD\textsubscript{i3d} (↓)  & FVD\textsubscript{cd} (↓) & MIoU (↑) \\
    \midrule
    V2V & 23.61 & 1201.26 & 123.05 & 36.13 \\
    SVS-GAN & \textbf{19.63}& \textbf{1103.69} & \textbf{96.18} & \textbf{45.08} \\
    \bottomrule
  \end{tabular}
  }}
  \caption{Quantitative results on the KITTI-360 dataset.}
  \label{tab:performance_comparison_KITTI}
\end{table}

\begin{figure}
  \centering
  \includegraphics[width=0.5\textwidth]{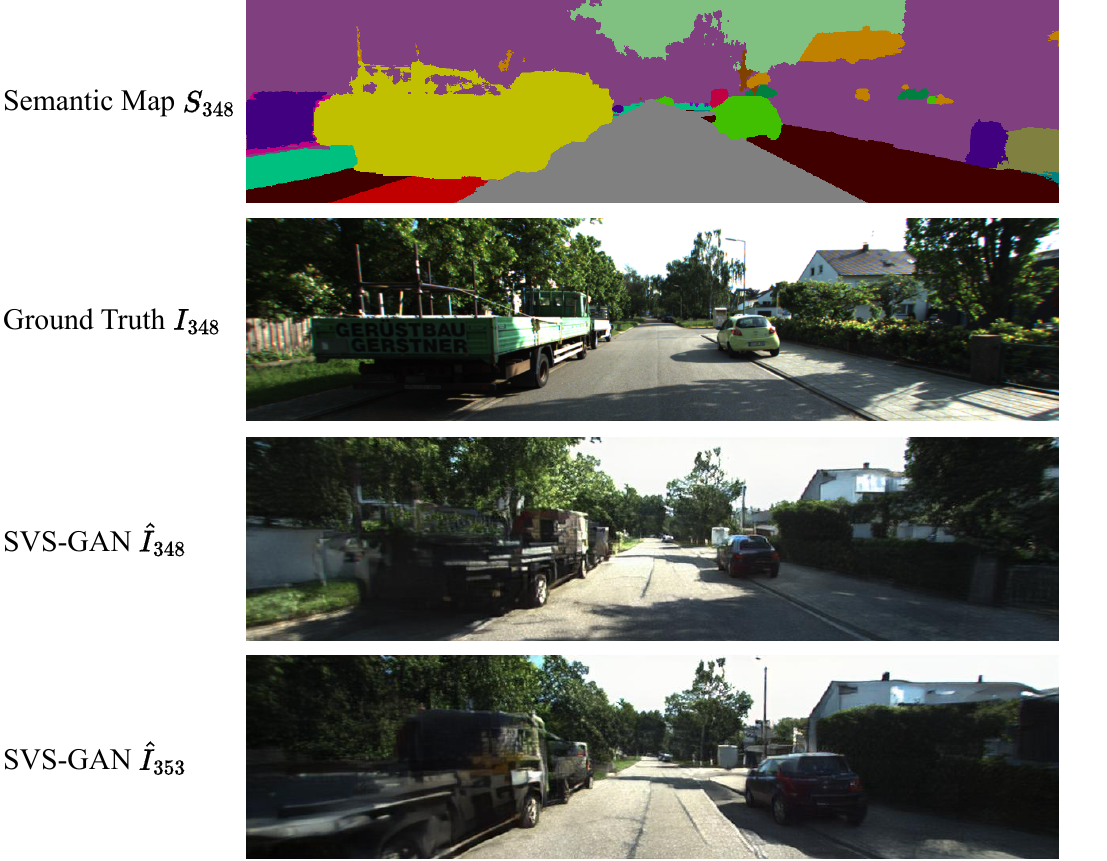}
  \caption{Our model demonstrates the ability to produce accurate results on long sequence lengths using the KITTI-360 dataset. The figure shows that even with rare classes like the truck, SVS-GAN maintains temporal consistency.}
  \label{fig:Results_KITTI360}
\end{figure}

Our model not only demonstrates accurate video generation but also achieves this with significantly fewer parameters and improved inference times, as evidenced by both quantitative and qualitative results. Figure \ref{fig:QualitativeCS} illustrates the capabilities of our framework by displaying the reference frame, the 10th, 20th, and last frames of each sequence alongside their semantic maps. In the first sequence, our model accurately preserves car details, closely aligning with the ground truth. Notably, our model uniquely reproduces the bicycle in the final frame, despite it not being visible in the reference image. This is a result of our discriminator prioritizing rare objects (the OASIS loss) and our generator's use of a triple-pyramid structure to balance global and local image aspects.

Conversely, WC-V2V struggles with the appearance of the road and lane colors, as seen in the first and second scenes. We can see an issue in V2V where the color of the van on the right changes between the 10th and 20th frames, indicating a lack of temporal coherence. WC-V2V performs better in rendering buildings due to its world view assumption, as buildings are static and give this approach an edge. However, obtaining a 3D world representation is not applicable in pure SVS frameworks.

The second scene in Figure \ref{fig:QualitativeCS} reveals a significant issue with WC-V2V: it assumes a static world representation, unaffected by time, leading to deteriorated results with dynamic objects in the scene. This limitation is illustrated where the people crossing the street behind the bus are missing from the WC-V2V output. This is because WC-V2V relies on the coloration of the 3D world representation when receiving new data. Since the 3D world is assumed to be constant, with only the movement of the ego-vehicle considered, WC-V2V fails in dynamic scenarios.

Additionally, in the second scene, V2V's reliance on a separate foreground network that does not depend on previous frames causes the flickering and inconsistent appearance of cars throughout the video. While this avoids the problem of complicated optical flow calculations for the movement of the cars, it introduces a larger issue where the appearance of the cars changes completely and is not coherent with the rest of the objects in the image. 

Similarly, V2V struggles with maintaining visual consistency for dynamically moving objects, such as the two people crossing the street. This example highlights our framework's capability in managing dynamic scenes and corner cases not represented in the training dataset. Accurately adhering to the semantic map in these scenarios is crucial, as they are essential for training robust frameworks that can generalize to critical real-world conditions often underrepresented in training datasets.

% The qualitative results are supported by quantitative evidence shown in Table \ref{tab:performance_comparison_CS}, comparing the frameworks on Cityscapes. Our model exhibits a slight decrease in FID, indicating superior frame-wise generation quality. A significant improvement in FVD reflects our model's enhanced temporal coherence, producing videos closely resembling real footage. Additionally, our MIoU is higher by more than 7 points, indicating that our generated videos more accurately follow the segmentation maps.

The qualitative results are supported by quantitative evidence presented in Table \ref{tab:performance_comparison_CS}, where our approach achieves best results throughout all metrics in comparison with the baselines on Cityscapes. Our model exhibits a decrease in FID, indicating better frame-wise visual quality and diversity. Moreover, significant improvements in both FVD\textsubscript{cd} and FVD\textsubscript{i3d} underscore our model's enhanced temporal coherence, with generated videos exhibiting realistic, smooth transitions. Additionally, our MIoU score exceeds others by over 4 points, demonstrating more accurate adherence to segmentation maps and confirming the model's effectiveness in maintaining visual and semantic integrity. Lastly, our framework not only significantly outperforms Fast-V2V\cite{fast_vi2vid} but also achieves this at a higher resolution compared to the $512 \times 256$ resolution of Fast-V2V, while both maintain a comparable frame rate of 25 FPS.

Further validation comes from the KITTI-360 dataset, as shown in Figure \ref{fig:Results_KITTI360}. Despite being trained on sequences with a maximum length of 30 frames, our model is capable of generating videos that exceed 1000 frames in length. This demonstrates the model's ability to generalize beyond its training constraints and maintain high-quality video generation over extended sequences.

The figure displays the generated 348th and 353rd frames, which include a rare class (truck). Our model successfully generates the truck with appropriate structure and detail, illustrating its capability to handle infrequent objects effectively. Additionally, it maintains temporal coherence between these frames which is critical for realistic video generation, ensuring that objects and their movements remain consistent over time.

Quantitative results presented in Table~\ref{tab:performance_comparison_KITTI} demonstrate that our SVS-GAN framework outperforms V2V on the KITTI-360 dataset. SVS-GAN not only achieves lower FID and FVD scores, indicating better image quality and spatial-temporal coherence, but it also shows a significant 8-point improvement in MIoU, further validating its enhanced capability to maintain semantic integrity in generated videos.

\subsection{Ablation Study}
To assess the impact of each modification, we performed an ablation study on the Cityscapes dataset. Due to long training times, we focused on the base model with a $512\times 256$ resolution, excluding spatial progressive learning. Consequently, the results differ from those presented in Table \ref{tab:performance_comparison_CS}.

Our results in Table \ref{tab:ablation_CS} illustrate the impact of each change where configuration A serves as our baseline. Adding the OASIS loss and the encoder/decoder discriminator in configuration B increased the MIoU by approximately 8 points, as the OASIS loss encourages the generator to produce results that the discriminator can easily segment. However, the image/video quality metrics did not show significant improvement, suggesting potential issues with the generator architecture. Key issues include the direct use of the semantic map as input along the reference image. Specifically, the SPADE paper\cite{SPADE} highlighted that this approach is sub-optimal for managing semantic maps, effectively 'washing away' semantic details essential for high-quality generation. Furthermore, the generated images are directly merged with the previously warped images without additional refinement, and the foreground module generating the cars fails to consider previous frames, resulting in flickering that severely undermines temporal coherence.

After modifying the generator architecture in framework C and solving the mentioned issues, we observed a decrease in FVD\textsubscript{i3d} by more than 30 points and a reduction in FID. This indicates that our new generator significantly enhances temporal consistency and video quality. Additionally, the MIoU increased, indicating that our generator receives better guidance with SPADE blocks. Interestingly, our new generator has significantly fewer parameters and produces better quality. This optimization was achieved through our design choices, such as employing fewer ResNet blocks and utilizing skip connections that facilitate better feature integration across the network. Additionally, by considering only one past frame for appearance , as opposed to V2V and WC-V2V which use two past frames, our model's complexity was reduced while maintaining high performance. 
\begin{table}
  \centering
  {\small{
  \begin{tabular}{@{}lccccc@{}}
    \toprule
     & FID (↓) & FVD\textsubscript{i3d} (↓)  & MIoU (↑) & $G$ Params  \\
    \midrule
    A & 46.88 & 110.87 & 58.56 & 411.3 M \\
    B: A+OASIS & 46.98 & 105.52&  66.66 & 411.3 M \\
    C: B+SPADE & \textbf{45.35} &  \textbf{72.38}& \textbf{68.47} & 105.9 M \\
    \bottomrule
  \end{tabular}
  }}
  \caption{Ablation study results on the Cityscapes dataset.}
  \label{tab:ablation_CS}
\end{table}

% \begin{table}
%   \centering
%   {\small{
%   \begin{tabular}{@{}lccccc@{}}
%     \toprule
%      & FID (↓) & FVD\textsubscript{i3d} (↓)  & FVD\textsubscript{cd} (↓) & MIoU (↑) & $G$ Params \\
%     \midrule
%     A & 62.68 & 110.87 & 71.75&53.25 & 411.3 M \\
%     B: A+OASIS & 61.04 & 105.52& 89.97& 59.80 & 411.3 M \\
%     C: B+SPADE & \textbf{58.47} & \textbf{72.38}& \textbf{58.08}& \textbf{62.43} & 105.9 M \\
%     \bottomrule
%   \end{tabular}
%   }}
%   \caption{Ablation study results on the Cityscapes dataset.}
%   \label{tab:ablation_CS}
% \end{table}

% \begin{table}
%   \centering
%   {\small{
%   \begin{tabular}{@{}lcccc@{}}
%     \toprule
%      & FID (↓) & FVD (↓) & MIoU (↑) & $G$ Params \\
%     \midrule
%     A & 33.97 & 127.56 & 46.64 & 411.3 M \\
%     B: A+OASIS & 35.05 & 138.62 & 52.49 & 411.3 M \\
%     C: B+SPADE & \textbf{31.56} & \textbf{72.11} & \textbf{53.97} & 105.9 M \\
%     \bottomrule
%   \end{tabular}
%   }}
%   \caption{Ablation study results on the Cityscapes dataset.}
%   \label{tab:ablation_CS}
% \end{table}
\begin{table}
  \centering
  {\small{
  \begin{tabular}{@{}lcccc@{}}
    \toprule
     & FID (↓) & FVD\textsubscript{i3d} (↓) & MIoU (↑) & $G$ Params (M) \\
    \midrule
    C & 45.35 & 72.38 & \textbf{68.47} & 105.9 \\
    D & 48.42 & 88.68 & 66.01 & 105.9 \\
    E & \textbf{43.81} & \textbf{70.54} & 67.97 & 153.0 \\
    \bottomrule
  \end{tabular}
  }}
  \caption{Comparing different generator architecture modifications.}
  \label{tab:ablation_bad}
\end{table}
We have performed multiple other experiments to validate our generator architecture. In our final architecture, we produce the optical flow from the two semantic maps and then warp the image before passing it to the image encoder. We tested feeding the image as is to the encoder and warping the features of the image that get fed to the decoder (Framework D). We found that our method performs better when looking at all the metrics as seen in Table \ref{tab:ablation_bad}. 

We have also tested in framework E the use of an extra SPADE block in each layer to take the style of the warped input image, similar to how we did with the semantic map. Although this modification led to slight improvements in both FVD\textsubscript{i3d} and FID compared to framework C, the MIoU slightly worsened. Furthermore, this approach resulted in almost a 50\% increase in the number of generator parameters, so we opted to go with our decided architecture.

\section{Conclusion}
In this paper, we defined Semantic Video Synthesis (SVS) and developed a GAN framework capable of generating realistic, time-coherent videos from corresponding semantic maps. Through a dedicated architecture and losses tailored for the SVS application, we achieved state-of-the-art results on the Cityscapes sequence dataset while maintaining a more efficient architecture. Additionally, we are the first to apply this task to the KITTI-360 dataset, successfully generating long sequences of realistic temporally coherent videos from the corresponding semantic maps.

Despite the success of our method in producing realistic, time-coherent results, we identified a limitation in maintaining world-consistency. Each generated frame relies only on the previous frame, without knowledge of the preceding frames. This can result in inconsistencies, such as a car appearing in different colors if it reappears after several frames. Addressing this issue without incorporating a 3D world representation as input is a critical area for future research. Furthermore, another significant limitation is the handling of optical flow, particularly in scenarios where a vehicle maneuvers around corners, such as turning right or left, or navigating roundabouts, where flow estimation becomes much more challenging. Additionally, managing domain shifts throughout extended video sequences presents a critical area for future exploration.

{\small
\bibliographystyle{ieee_fullname}
\balance
\bibliography{egbib}
}

\end{document}

\title{SVS-GAN: Leveraging GANs for Semantic Video Synthesis \\ Supplementary Material}  % **** Enter the paper title here

\maketitle
\thispagestyle{empty}
\appendix

%%%%%%%%% BODY TEXT - ENTER YOUR RESPONSE BELOW
\section{Losses}

Our architecture is optimized using an extensive set of loss functions designed to refine different aspects of the generated output. Throughout the following equations, \(x\) denotes the real video frame, and \(s\) represents the input semantic label map. Additionally, \(\hat{x}\) is defined as \(\hat{x} = G(x_p, s_p, s)\), where \(x_p\) is the previously generated image, and \(s_p\) and \(s\) are the past and current semantic maps, respectively. The individual losses and their formulations are as follows:

\begin{itemize}
\item \textbf{OASIS Adversarial Loss for $D_I$:} As \cite{OASIS}, to ensure that the generator synthesizes images that align with the input semantic label maps, we use a discriminator that can perform both semantic segmentation and fake/real detection. This is achieved by casting the discriminator task as a multi-class semantic segmentation problem. The loss is defined as:

\begin{equation}
\begin{aligned}
\mathcal{L}_{D_I} = & - \mathbb{E}_{x} \left[ \sum_{c=1}^{N} \alpha_{c} \sum_{i,j}^{H \times W} s_{i,j,c} \log D_I(x)_{i,j,c} \right] \\
& - \mathbb{E}_{\hat{x}} \left[ \sum_{i,j}^{H \times W} \log D_I(\hat{x})_{i,j,c=N+1} \right]
\end{aligned}
\label{DIDloss}
\end{equation}

where $N$ is the number of real classes, $\alpha_c$ is the weight for class $c$, $s_{i,j,c}$ is the label at position $(i,j)$ for class $c$, $D_I(x)_{i,j,c}$ is the discriminator output at position $(i,j)$ for class $c$, and $D_I(\hat{x})_{i,j,c=N+1}$ is the discriminator output at position $(i,j)$ for the fake class $N+1$. Here, $H \times W$ are the height and width of the input data.

To balance the contributions of each class, we weight each class by its inverse per-pixel frequency, giving more importance to rare classes and encouraging the generator to synthesize less-represented classes adequately.

\item \textbf{OASIS Adversarial Loss for $G$\cite{OASIS}:} The objective of this loss is to make sure that the images generated by $G$ are both realistic and semantically consistent with the input label maps. This is achieved by training the generator to maximize the likelihood that the discriminator $D_I$ will classify the generated images as belonging to the correct semantic classes.

\begin{equation}
\begin{aligned}
\mathcal{L}_{G_I} = & - \mathbb{E}_{\hat{x}}\left[ \sum_{c=1}^{N} \alpha_{c} \sum_{i,j}^{H \times W} s_{i,j,c} \log D_I(\hat{x})_{i,j,c} \right]
\end{aligned}
\label{DIGloss}
\end{equation}

\item \textbf{Adversarial Loss for $D_V$\cite{vid2vid,WC-vid2vid}:} Utilized with the video discriminator $D_V$, this loss encourages temporal coherence:
\begin{equation}
\begin{aligned}
\mathcal{L}_{adv} = \mathbb{E}_{\hat{v}}[\log D_V(\hat{v})] + \mathbb{E}_v[\log (1 - D_V(v))]\mathcal{
}\end{aligned}
\label{ADVloss}
\end{equation}

In this formulation, $v$ represents real video sequences and $\hat{v}$ represents generated video sequences. The loss is designed to encourage the production of videos that are not only realistic in appearance but also temporally coherent, ensuring smooth transitions and consistent motion across frames.

    \item \textbf{VGG Loss\cite{VGGGan}:} This loss ensures that the generated images are perceptually similar to the real images by utilizing features extracted from multiple layers of a VGG network\cite{VGG}. The VGG loss measures the difference between the feature representations of the generated image and the real image:
\begin{equation}
\begin{aligned}
\mathcal{L}_{VGG} = & \mathbb{E}_{(x, \hat{x})} \left[ \sum_{l=1}^{L} \beta_{l} \left\| \phi_l^{VGG}(\hat{x}) - \phi_l^{VGG}(x) \right\|_1 \right]
\end{aligned}
\label{VGGloss}
\end{equation}

Here, \(\phi_l^{VGG}\) denotes the feature maps from layer \(l\) of the VGG network, and \(L\) is the total number of layers used.

\item \textbf{Feature Matching Loss\cite{fmloss}:} This loss ensures that the feature representations of generated images are closely aligned with those of real images, by comparing outputs from specific layers of the discriminators $D_I$ and $D_V$. The feature matching loss is computed as the L1-norm difference between the features of the generated image and the real image across multiple layers:

\begin{equation}
\begin{aligned}
\mathcal{L}_{FM} = & \mathbb{E}_{(x, \hat{x})} \left[ \sum_{l=1}^{L} \alpha_l \left\| \phi_l^{D}(\hat{x}) - \phi_l^{D}(x) \right\|_1 \right]
\end{aligned}
\label{FMLoss}
\end{equation}

Here, \(\phi_l^{D}\) represents the output from the \(l\)-th layer of the discriminators ($D_I$ or $D_V$), where \(L\) denotes the total number of layers evaluated.

\item \textbf{Flow and Warping Loss:} As \cite{vid2vid, WC-vid2vid}, this loss enhances the network's optical flow predictions, using ground truth flows $OF_{i-1 \rightarrow i}$ derived via FlowNet2\cite{flownet}. It combines the flow accuracy and image warping errors:
\begin{equation}
\begin{aligned}
L_{Flow} &= \lambda_{OF}\| OF_{i-1 \rightarrow i} - \widehat{OF}_{i-1 \rightarrow i} \|_1 \\
& + \lambda_W \| WI_{i-1} - I_i \|_1
\end{aligned}
\end{equation}

In this equation, $\widehat{OF}_{i-1 \rightarrow i}$ indicates the predicted flow from $G$, used to warp the previous image $I_{i-1}$ into $WI_{i-1}$ by shifting pixels based on their motion vectors. The term $\lambda_{OF}$ and $\lambda_{W}$ balances the importance of flow accuracy and warping discrepancies relative to the total loss.

\end{itemize}

The final loss function for our generator model is composed of several components, weighted appropriately based on insights derived from prior frameworks. The composition of the generator loss function is detailed as follows:
\begin{equation}
\begin{aligned}
\mathcal{L}_{G} &= \mathcal{L}_{G_I} + \mathcal{L}_{adv} + L_{Flow} \\
&+ \lambda_{VGG} \mathcal{L}_{VGG} + \lambda_{FM} \mathcal{L}_{FM} 
\end{aligned}
\end{equation}

These losses ensure that our architecture not only generates realistic and temporally coherent video sequences but also adheres closely to the given semantic maps.

\section{Metrics}
For the evaluation of our SVS-GAN, we employ several key metrics that quantify both the visual quality and the semantic accuracy of the generated videos:

\begin{itemize}
    \item \textbf{Fréchet Inception Distance (FID)\cite{FID}:} This metric measures the distance between feature vectors calculated for real and generated images. The vectors are obtained by a deep convolutional network, Inception-v3\cite{inception}, trained on ImageNet\cite{imagenet}. A lower FID score indicates that the generated images are closer to the real images in terms of their content and style, suggesting better model performance.

    % \item \textbf{Fréchet Video Distance (FVD)\cite{FVD}:} Designed to evaluate the quality of generated videos, FVD extends the concept of FID to the temporal domain. It compares the statistical distribution of features extracted from real video clips and synthetic video clips generated by the model. A lower FVD value shows higher similarity between real and synthesized videos, indicating superior temporal coherence and visual quality.

    \item \textbf{Fréchet Video Distance (FVD):} The FVD metric evaluates the quality of generated videos by measuring the Fréchet distance between the feature distributions of real and synthetic video clips. We employ two variants: FVD\textsubscript{i3d} and FVD\textsubscript{cd} as implemented in \cite{FVDcd}. FVD\textsubscript{i3d} uses features extracted via the I3D network, trained on the Kinetics-400 dataset\cite{i3d}, focusing mainly on frame appearance. In contrast, FVD\textsubscript{cd}, or content debiased FVD, utilizes features from the VideoMAE-v2 model\cite{videomae}, which is trained in a self-supervised manner on diverse video data. Lower values in both metrics suggest higher similarity and quality, with FVD\textsubscript{cd} providing a more precise evaluation of spatial and temporal coherence. Moreover, \cite{FVDcd} observed that significant variations between FVD\textsubscript{i3d} and FVD\textsubscript{cd} may occur, depending on the dataset.

    \item \textbf{Mean Intersection over Union (MIoU):} To assess the semantic accuracy of the generated video frames, we compute MIoU using the DRN-D-105\cite{drn} network, a deep residual network tailored for semantic segmentation. MIoU measures the pixel-wise overlap between the predicted semantic labels and the ground truth across the semantic classes. Higher MIoU values indicate better semantic segmentation performance, reflecting the model’s ability to understand and replicate the scene's contextual details.
\end{itemize}

\begin{figure*}
\centering
\includegraphics[width=1.0\textwidth]{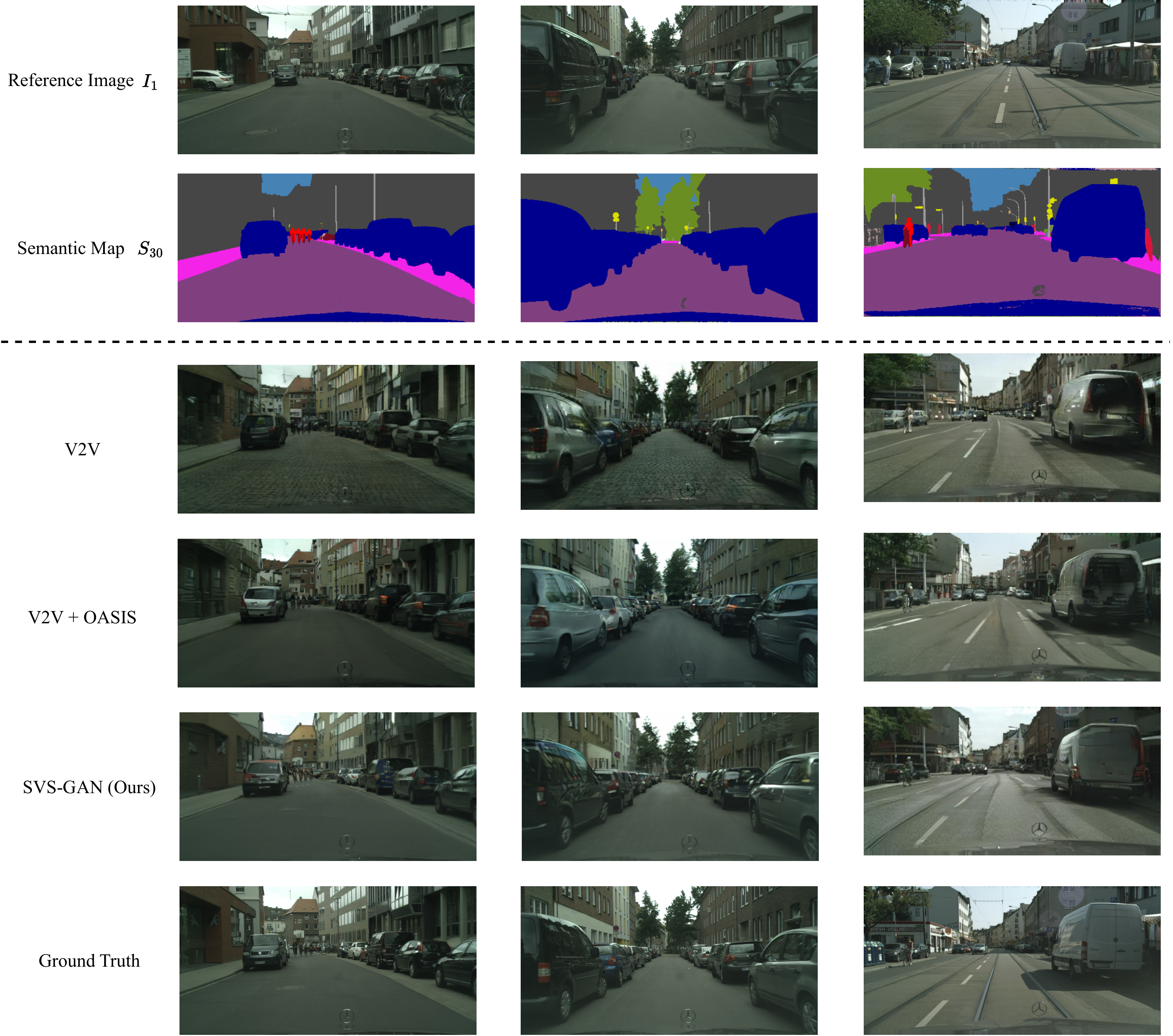}
\caption{Qualitative Ablation Study: We present a comparative analysis involving SVS-GAN, V2V, and V2V with OASIS loss in this figure. Displayed are the reference image and the semantic map for the 30th frame, alongside the synthesized 30th frame image and its corresponding ground truth. The incorporation of the OASIS loss appears to enhance the model's attention to fine details, such as street signs and bicycles. Moreover, our SVS-GAN exhibits better performance in detail preservation and spatial coherence due to our new generator.}
\end{figure*}

%%%%%%%%% REFERENCES
{\small
\bibliographystyle{ieee_fullname}
\bibliography{egbib}
}